\documentclass[letterpaper, 10 pt, journal, twoside]{IEEEtran}

\IEEEoverridecommandlockouts                              

\usepackage{url}
\usepackage{graphics} 
\usepackage{epsfig} 
\usepackage{mathptmx} 
\usepackage{times} 
\usepackage{amsmath} 
\usepackage{amssymb}  
\usepackage{subfigure}
\usepackage[T1]{fontenc}
\usepackage[vlined,ruled]{algorithm2e}
\usepackage{booktabs}
\usepackage{multirow}

\usepackage{graphicx}
\usepackage{float}

\usepackage{cite}
\usepackage{algpseudocode}
\usepackage[colorlinks,linkcolor=blue,anchorcolor=blue,citecolor=green]{hyperref} 
\usepackage[export]{adjustbox}
\usepackage{flushend}
\usepackage{rotating}
\usepackage[dvipsnames]{xcolor}

\newcommand{\greenup}{\textcolor{Green}{$\uparrow$}}
\newcommand{\reddown}{\textcolor{Red}{$\downarrow$}}

\begin{document}

\title{\LARGE \bf
STTracker: Spatio-Temporal Tracker for 3D Single Object Tracking
}

\author{Yubo Cui, Zhiheng Li, Zheng Fang$^*$
\thanks{Manuscript received: February, 11, 2023; Revised May, 12, 2023; Accepted June, 12, 2023.}
\thanks{ This paper was recommended for publication by Editor Cesar Cadena upon evaluation of the Associate Editor and Reviewers’ comments.
This work was supported in part by the National Natural Science Foundation of China
under Grants 62073066 and U20A20197, in part
by the Fundamental Research Funds for the Central Universities under Grant N2226001, and in part by 111 Project under Grant B16009. (Corresponding author: Zheng Fang.)}
\thanks{Yubo Cui, Zhiheng Li, Zheng Fang are with Faculty of Robot Science and Engineering, Northeastern University, Shenyang 110819, China, and also with the National Frontiers Science Center for Industrial Intelligence and Systems Optimization, Shenyang 110819, China (e-mail: ybcui21@stumail.neu.edu.cn, zhihengli@stumail.neu.edu.cn, fangzheng@mail.neu.edu.cn)}
\thanks{Zheng Fang is also with the Key Laboratory of Data Analytics and Optimization for Smart Industry Northeastern University), Ministry of Education, Shenyang 110819, China.}
\thanks{Digital Object Identifier (DOI): 10.1109/LRA.2023.3290524.}
}

\markboth{IEEE Robotics and Automation Letters. Preprint Version. Accepted June, 2023}
{Cui \MakeLowercase{\textit{et al.}}: STTracker: Spatio-Temporal Tracker for 3D Single Object Tracking} 

\maketitle

\begin{abstract}
3D single object tracking with point clouds is a critical task in 3D computer vision. Previous methods usually input the last two frames and use the predicted box to get the template point cloud in previous frame and the search area point cloud in the current frame respectively, then use similarity-based or motion-based methods to predict the current box. Although these methods achieved good tracking performance, they ignore the historical information of the target, which is important for tracking.
In this paper, compared to inputting two frames of point clouds, we input multi-frame of point clouds to encode the spatio-temporal information of the target and learn the motion information of the target implicitly, which could build the correlations among different frames to track the target in the current frame efficiently. 
Meanwhile, rather than directly using the point feature for feature fusion, we first crop the point cloud features into many patches and then use sparse attention mechanism to encode the patch-level similarity and finally fuse the multi-frame features. Extensive experiments show that our method achieves competitive results on challenging large-scale benchmarks (62.6\% in KITTI and 49.66\% in NuScenes).
\end{abstract}

\section{INTRODUCTION}
\label{sec:intro}
Single object tracking with point clouds is one of the most important tasks in 3D computer vision. Given the 3D box of the target in the initial frame, single object tracking requires the tracker to make successive predictions of the given target in subsequent frames to obtain the target's 3D pose, which could provide useful information for downstream tasks, such as path planning in autonomous following robots.

Currently, most previous methods~\cite{SC3D,P2B,BAT,PTT,lttr,v2b} use similarity computation to match the current frame point cloud with the template point cloud of the tracking target, and then find the target in the current frame point cloud. Meanwhile, since the first frame point cloud is the most accurate and has strong prior information, they usually update the template point cloud by fusing the predicted target point cloud of the previous frame with the initial frame target point cloud to achieve the best tracking results. However, the performance of this similarity-based matching paradigm is often limited due to the sparsity and disorder of point clouds. 
Recently, $M^2$-Tracker~\cite{beyond} proposed a motion-based tracking paradigm, suggesting that regressing the relative target motion from the two consecutive point cloud frames could be more suitable for the point cloud tracking task than similarity-based matching. 
By inputting the last two frames of point cloud and the predicted box of the previous frame, they first segmented the two point clouds to obtain the foreground point cloud, i.e., the approximate target point cloud.
Then, by regressing the offset between the two foreground point clouds, a coarse current box prediction is obtained based on the previous predicted 3D box. Finally, they fused the target point cloud in two frames and refined the coarse box to get a more accurate box.

\begin{figure}
	\centering
	\includegraphics[width=\linewidth]{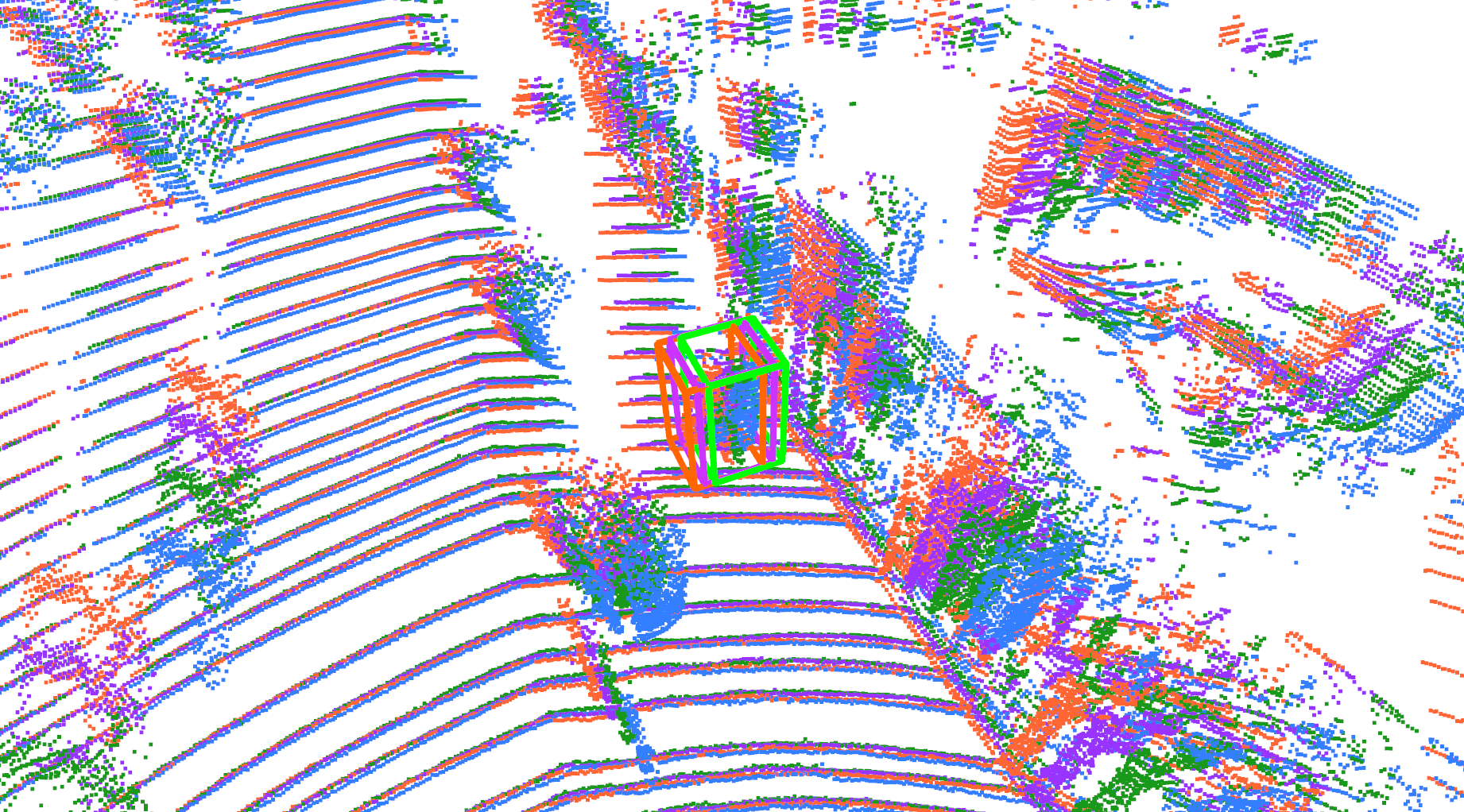}
	\caption{Multi-frame point clouds input. Our input includes $N$ frames point clouds and the past $N-1$ frame 3D boxes of target. Different colors represent different timestamps.}
	\label{fig:mf_points}
\end{figure}

However, no matter the similarity-based matching or motion-based estimation, they both only input two consecutive point cloud frames and ignore the earlier historical information of the target, which is also important for the tracking task. For example, if the target is taking a turn in recent frames, this long-time global motion information can be used to constrain the angle prediction in the current frame, while this motion information is difficult to be detected in only two local frames. Also, all previous similarity-based algorithms complement the template point cloud with the target information from the first frame. However, this skip-completion ignores the successive spatio-temporal information of the target in the historical frames and only superimposes the aligned point clouds, thus also not fully utilizing the spatio-temporal information during tracking.

To address the above issues, in this paper, we propose a point cloud tracking algorithm based on spatio-temporal information, termed STTracker. Different from previous works, we input the point clouds of the past $N-1$ frames and their corresponding 3D boxes of the target, as well as the point cloud of current frame to predict the current 3D box, as shown in Fig.~\ref{fig:mf_points}. Meanwhile, the input can be of any length and any frame, such as $[t, t-1, t-2, t-3]$ or $[t, t-2, t-4]$, etc. Therefore, using two consecutive frames of point clouds as input in previous methods can be considered as one of our input modes. After getting the multi-frame input, we propose a similarity-based spatio-temporal fusion module to build correlations between multi-frame point clouds and fuse the historical information into the current frame features for prediction. Given the previous 3D boxes of the target, the fusion module could learn the motion information implicitly. Furthermore, to reduce the computational effort of the fusion module and speed up the training and inference speed, we use a sparse patched-based attention mechanism for multi-frame feature fusion. 
Our method proves that by learning the spatio-temporal information of the target from multiple historical frames, the similarity-based matching paradigm could break the limitations and track the target with point clouds effectively. 
Compared to $M^2$-Tracker~\cite{beyond} which only learns the short motion information between two frames, our method not only fully utilizes the long spation-temporal information brought by multiple frames to implicitly learn motion information, but also learns the appearance similarity information between multiple frames to better locate the target position.
Comprehensive evaluation results show that our STTracker achieves competitive results on KITTI~\cite{KITTI} and NuScenes~\cite{NuScenes} datasets. 

Overall, our contributions are as follows:
\begin{itemize}	
    \item We propose a spatio-temporal learning framework that introduces multiple frames into 3D single object tracking.
    
    \item We propose a novel multi-frame features fusion method to implicitly learn the motion information of the target and build correlation among multiple frames.
    
	\item Experiments on KITTI and NuScenes datasets show that our STTracker achieves promising performance, and extensive ablation studies also verify the effectiveness of our method.
\end{itemize}

The rest of this paper is organized as follows. In Sec. \ref{sec:related},
we discuss the related work. Sec. \ref{sec:methodology} describes the proposed STTracker. In Sec. \ref{sec:experiments} we first compare our methods with previous methods in KITTI and NuScenes datasets, and then conduct ablation studies on each module of our methods. We finally conclude in Sec. \ref{sec:conclusions}.

\section{RELATED WORK}\label{sec:related}
\subsection{3D Single Object Tracking}
3D single object tracking with point cloud has developed fast in recent years. SC3D~\cite{SC3D} compares template and search point clouds with cosine similarity and selects the highest score one to track.
P2B~\cite{P2B} proposes an augmentation module to fuse the point features with the point-wise cosine similarity, and then takes VoteNet~\cite{votenet} to have an accurate 3D box. 
Following this pipeline, PTT~\cite{PTT}, BAT~\cite{BAT}, LTTR~\cite{lttr}, V2B~\cite{v2b}, PTTR~\cite{pttr} and SMAT~\cite{smat} also take the two-branch siamese architecture and similarity-based matching paradigm. By enhancing the point features~\cite{PTT,BAT,v2b}, computing the similarity with transformer~\cite{lttr,pttr,smat}, the similarity-based matching pipeline makes great progress in 3D single object tracking task. 
PCET~\cite{pcet} proposes two modules to extract discriminative features and improve the robustness to sparse point clouds and respectively.
Different from these similarity-based methods,
$M^2$-tracker~\cite{beyond} points out that the motion-based tracking paradigm may be more suitable for 3D SOT than similarity matching. They first segment the foreground points to find the target points and then regress the relative target motion between the two frame points to get a coarse 3D box. Finally, they aggregate the target from two successive frames by using the predicted motion state and refine the coarse 3D box to get a better prediction. 
Moreover, STDA~\cite{stda} proposes a temporal motion model to learn the spatio-temporal information to track object by predicting the state and variance of the target.
However, their method depends on the detector and could not track the object in end-to-end manner.
In this paper, different from STDA~\cite{stda}, we propose an end-to-end network to directly learn the spatio-temporal information from multi-frame data.

\begin{figure}[t]
	\centering
	\includegraphics[width=\linewidth]{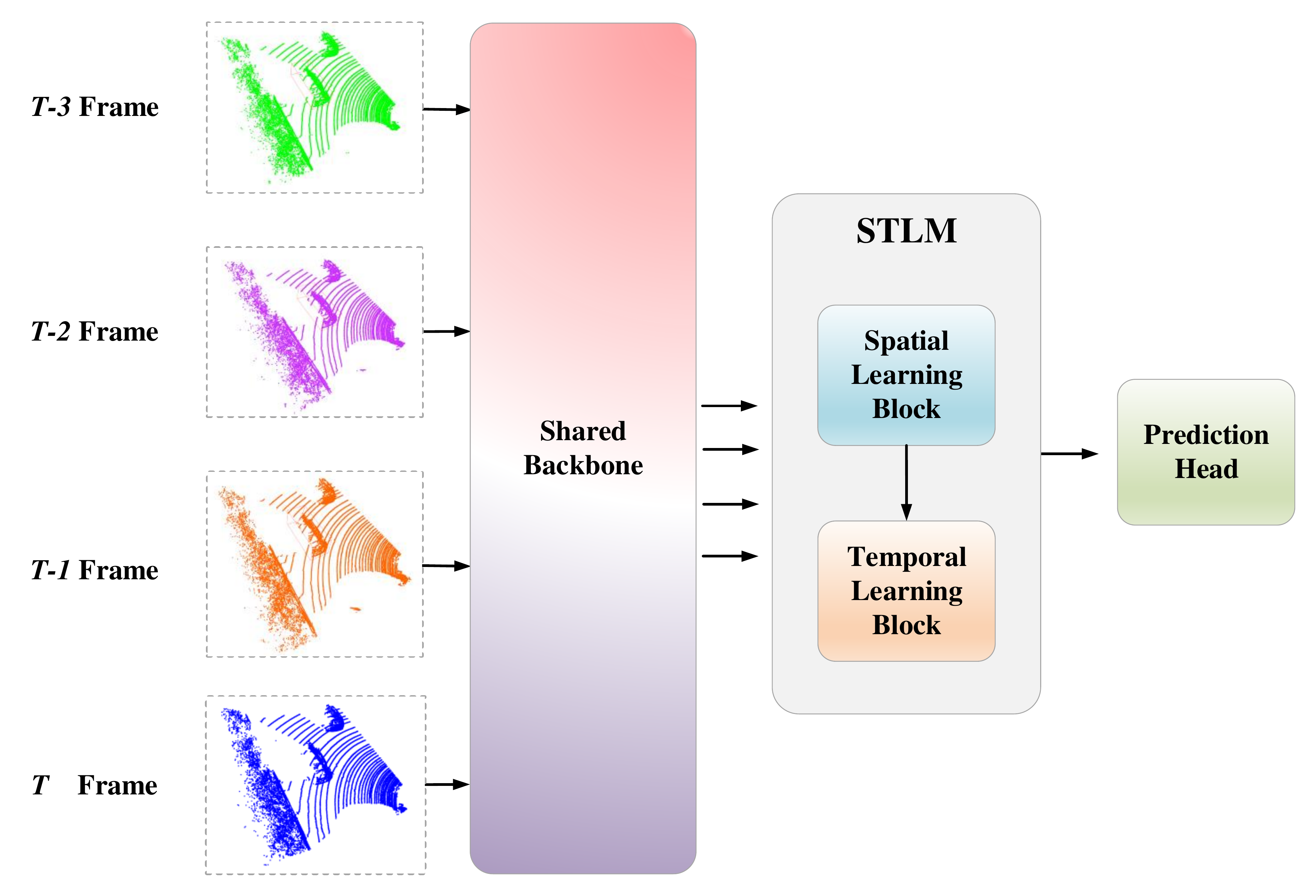}
	\vspace{-0.25in}
	\caption{Architecture of the proposed STTracker. Given $N$ point cloud and corresponding $N-1$ 3D boxes, we \textbf{first} use a shared backbone to extract features from point clouds. \textbf{Second}, we input $N$ features with $N-1$ 3D boxes into our spatio-temporal fusion module to learn spatio-temporal information. \textbf{Finally}, we use a center-based regression to predict the current box.}\label{fig:overall}
\end{figure}

\begin{figure*}[t]
	\centering
	\includegraphics[width=0.9\linewidth]{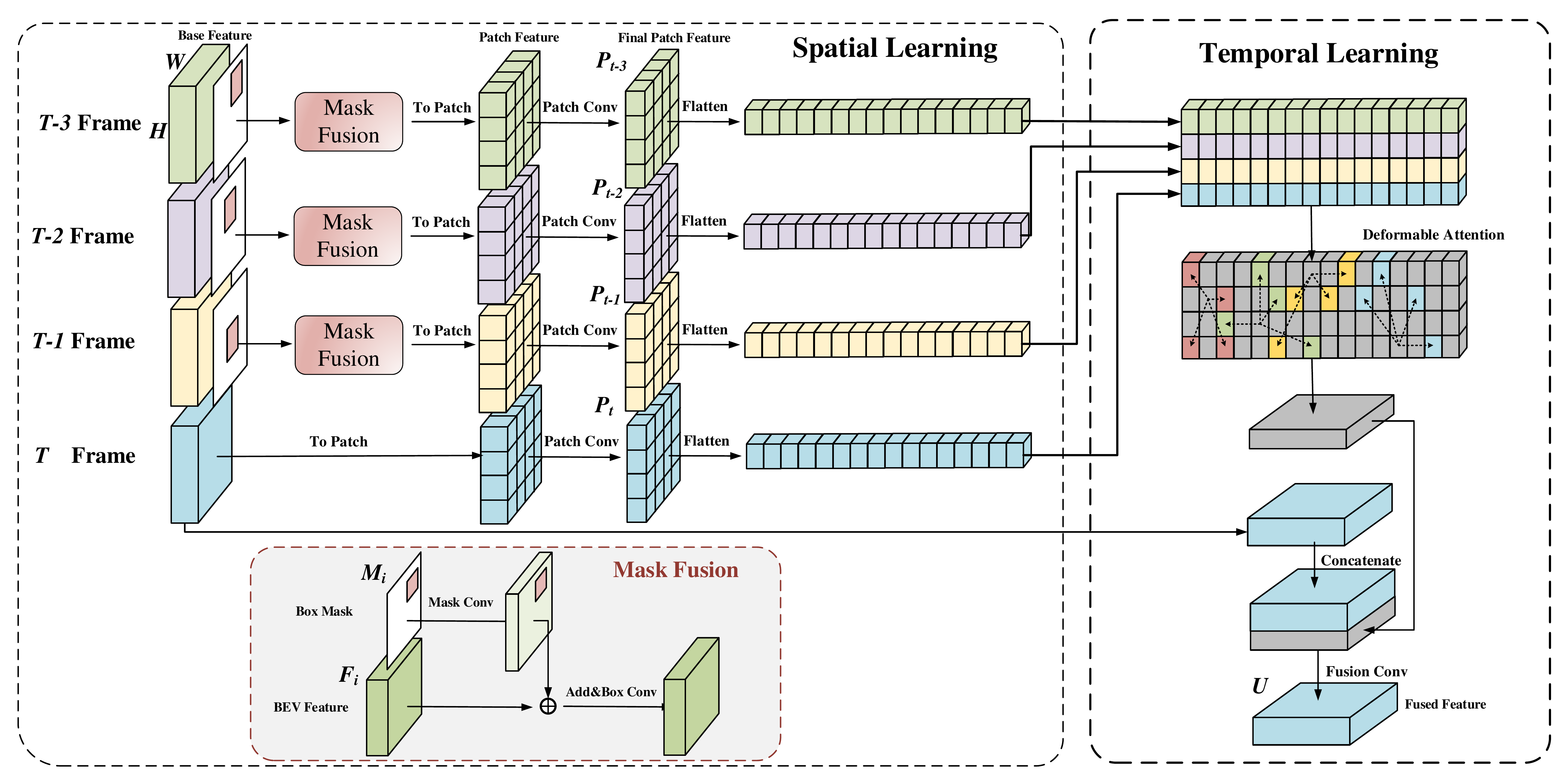}
	\vspace{-0.1in}
	\caption{Illustration of the proposed STLM. The STLM includes two components, spatial learning block and temporal learning block respectively. The first block learns the spatial information for each frame feature, and the second block learns the temporal information from all frame features.}\label{fig:stneck}
\end{figure*}

\subsection{Spatio-Temporal Learning}
Learning spatio-temporal information across multiple frames has been exploited for 3D vision tasks, such as 3D object detection and point cloud prediction.
Faf~\cite{faf} jointly conducts object detection, tracking, and motion forecasting together by inputting multiple point cloud frames and designs two schemes for temporal fusion.
StarNet~\cite{StarNet} uses the previous detection results as prior to improve the current detection. STINet~\cite{STINet} inputs multiple frames to extract features and temporal proposals to detect in current frame and predict future trajectories simultaneously. MVFuseNet~\cite{MVFuseNet} designs multi-view temporal fusion of LiDAR in RV and BEV for current detection and motion forecasting. 3DSTCN~\cite{st3dcnn} projects past $N$ frame point clouds into 2D range images and apply a U-Net-like spatio-temporal 3D CNN to obtain the future 3D point cloud predictions. MGTANet~\cite{mgtanet} designs short-term feature extraction and long-term feature enhancement to learn spatial-temporal information.
SpOT~\cite{spot} represents tracked objects as sequences and proposes a sequence-level 4D refinement network. PF-Tracker~\cite{pf_track} proposes a multi-camera 3D MOT framework that adopts the “tracking by attention” pipeline.

\section{METHODOLOGY}
\label{sec:methodology}
\subsection{Overall Architecture}
Given the current point cloud, and previous $N-1$ frames point clouds with their corresponding 3D bounding boxes, we aim at estimating the current target 3D box, which could be represented as $(x,y,z,w,l,h,\theta)$, where $(x,y,z)$ is the center, $(w,l,h)$ is the size and $\theta$ is the orientation of the box respectively. Meanwhile, following the assumption~\cite{P2B} that the size of the target object is known through the first frame, we only need to estimate $(x,y,z,\theta)$. 

As shown in Fig.~\ref{fig:overall}, our proposed STTracker (\textbf{S}patial \textbf{T}emporal \textbf{T}racker) is a one-stage network that has a simple pipeline. We first input all $N$ frames of point cloud into a shared backbone to extract per-frame point features. 
Unlike previous works~\cite{P2B,PTT,lttr,BAT} which only input the points within the predicted 3D box in the template branch, our $N-1$ previous frames of point cloud adopt the same input size as the current frame, as shown in Fig.~\ref{fig:overall}.
Meanwhile, we add the timestamps for all points to construct time-aware point cloud $\mathcal{P}_t =\{x,y,z,t\}$, and use dynamic pillar~\cite{dynamic} to extract basic features for its fast speed. We refer the readers to paper~\cite{dynamic} for more details.
Second, we input all $N$ extracted features and previous $N-1$ 3D boxes to our fusion module to learn spatio-temporal information. Finally, by using center-based prediction, we predict the 3D box of the target in current frame. We will introduce the details in the following subsections.

\subsection{Spatio-Temporal Learning}\label{sec:stlearning}
After per-frame feature extraction, we have $N$ frames of point cloud features and their sizes are all $W\times H \times C_1$, where $N-1$ are from previous frame point clouds and the last one is from the current point cloud, $C_1$ is the feature channel. Meanwhile, 
we multiply the voxel size with the final downsampling rate of our network to get the final grid size. Based on the point cloud range and the grid size, we generate a corresponding empty BEV (Bird's Eye View) grid map ($H\times W\times 1$). We assign the mask value of each grid as 1 if the center of its center is in the BEV box, otherwise as 0.
Therefore, we could obtain $N-1$ box masks.
Our goal is to extract useful spatio-temporal relationships to guide current prediction. The simplest approach is concatenating them together and applying 2D convolutional block to directly extract the spatio-temporal features. However, we argue that this approach could not learn the information efficiently due to the misalignment among features at different timestamps. Because of the motion from the LiDAR sensor or the target itself, 
the same position in the feature maps from different timestamp features represents different point clouds. 
Although sometimes the ego-motion of the LiDAR sensor is available, the motion of the target is always unknown. Therefore, simple concatenation would lead to ambiguity of features~\cite{mgtanet}. Another approach is applying 3D convolutional block~\cite{st3dcnn} to the concatenated features. However, the 3D convolutional block would incur huge computation cost.

To align the features from different timestamps and learn the spatio-temporal information of the target, we proposed an attention-based feature fusion module, as shown in Fig.~\ref{fig:stneck}, termed STLM (\textbf{S}patio-\textbf{T}emporal \textbf{L}earning \textbf{M}odule). STLM adopts a similarity-based matching to fuse different timestamp features, and could be divided into spatial learning block and temporal learning block respectively.

\textbf{Spatial Learning Block.}
The spatial learning block aims to learn the spatial information for each frame feature, thus it only involves single-frame features. Different from previous similarity-based works~\cite{P2B,PTT,BAT,v2b,lttr}, we input previous point clouds not only including the points within the 3D box but also including the points out of the 3D box, as the same as the current search point cloud. Therefore, to distinguish the foreground points from input, we need a BEV mask to represent the box's spatial location. 
We propose the $MaskFusion$ to incorporate the box mask into the extracted features. As shown in Fig.~\ref{fig:stneck}, we first apply a Conv2D layer named \textit{MaskConv} to project the mask into features, then add the mask feature with the BEV features and further apply a Conv2D layer named \textit{BoxConv} to further extract the foreground features with channel $C_2$. Compared to the methods which only extract the feature from points within the 3D box, we could keep and extract much more texture information. This operation could be formulated as follows:
\begin{equation}
	 \hat{F_i}= {\rm BoxConv}({\rm MaskConv}(M_i)+ F_i)\label{equ:attention}
\end{equation}
where $i \in \{t-1,...,t-N\}$ and $F_i, M_i$ denote the point feature and box mask for $i$-th frame, respectively. Meanwhile, since we need to compute the similarity among $N$ frames of features in the following temporal learning block, the $W \times H$ size would incur high computation cost. 
Therefore, following previous vision transformer works~\cite{VIT,pvt,TNT,swin}, we also divide per-frame feature map into many non-overlapping local patch grids.
Specially, we set the patch size to $R\times R$ and crop per-frame $W \times H \times C_2$ features to $S$ patches with size of $R \times R \times C_2$, then apply a Conv2D layer named \textit{PatchConv} to extract the per-patch features with channel $C_3$. Finally, we flatten the $R \times R\times C_3$ patch features to $S \times C_3$, where $S=R\times R$. Denoting the patch transformation as $\phi$, this procedure is represented as follows:
\begin{equation}
P_i = {\rm PatchConv}(\phi[\hat{F_i}])
\end{equation}
where $i \in \{t,...,t-N\}$ and $P_i$ refers to the final patch features. The patch transformation could not only reduce the computation cost, but also provide a larger receptive field for the following similarity-based matching.

Overall, we apply the proposed $MaskFusion$ and patch transformation $\phi$ for all $N-1$ previous features, while we only apply the patch transformation $\phi$ for current frame feature since we do not have the 3D box of current frame.

\textbf{Temporal Learning Block.}
Given $N$ patch features $P_i$ from different timestamps, the temporal learning block uses a sparse attention-based paradigm to fuse them and finally outputs the feature including the spatio-temporal information of the target. In particular, we first concatenate these per-frame patch features to have the fused spatial-temporal feature $G$ of size $N \times S \times C_3$, where the horizontal axis represents space and the vertical axis represents time. Then, inspired by the attention mechanism~\cite{Transformer}, we apply the deformable attention~\cite{DeformableDETR} to align the different timestamp patch features by themselves. 
Specially, we first use two linear layers for each query patch feature to generate sampling offsets $\Delta G$ and attention weights $A$ respectively. Based on the location of query patch itself and the output sampling offsets, we can sample reference patch features from the feature $G$ by bilinear interpolation. Finally, the original patch feature could be aggregated with the reference features with their corresponding weights. The procedure could be formulated as follows:
\begin{equation}~\label{equ:offset}
\Delta G = {\rm MLP_o}(G)
\end{equation}
\begin{equation}~\label{equ:sim}
A = {\rm MLP_s}(G)
\end{equation}
\begin{equation}~\label{equ:value}
V_k = S(G, g+\Delta g_k)
\end{equation}
\begin{equation}~\label{equ:head}
\hat{G} = \sum_{l=1}^LW_l\left(\sum_{k=1}^KA_{lk}\cdot V_k\right)
\end{equation}
where $S(\cdot)$ is the bilinear sampling, $K$ is the number of predicted offset for each grid, $L$ is the number of attention heads. In Equ.~\ref{equ:offset} and Equ.~\ref{equ:sim}, we use two MLP layers to predict $K$ offsets $\Delta g_k$ and corresponding similarity scores $A_k$ for each feature grid respectively. Then, in Equ.~\ref{equ:value}, we add the offsets $\Delta g_k$ to the grid coordinate $g$ to get new grid coordinate and use bilinear sampling $S(\cdot)$ to sample the feature at $g_k+\Delta g_k$ from $G$. Moreover, in Equ.~\ref{equ:head}, the sampled features $V_k$ are multiplied with the similarity scores $A_k$. Finally, following the multi-head mechanism,  we use $W_l$ to project each head feature back and sum them up.
We refer the reader to paper~\cite{DeformableDETR} for more details.

Finally, we reshape the fused feature $\hat{G}$ back into the size of $W \times H \times C_4$ and concatenate it with the original current frame feature $W \times H \times C_1$ to strengthen current frame features. The final fused feature $U$ is generated as follows:
\begin{equation}
U ={\rm Conv2D}({\rm cat}[\hat{G},F_t])
\end{equation}
By using the sparse deformable attention~\cite{DeformableDETR}, each patch could find its corresponding region to align based on the similarity. Meanwhile, the sparsity also avoids all-to-all similarity matching and further limits the computation cost.

\begin{figure}[t]
	\centering
	\subfigure[]{
		\includegraphics[width=0.25\linewidth]{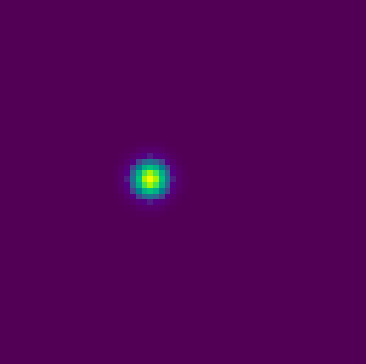}
	}
	\quad
	\subfigure[]{
	\includegraphics[width=0.25\linewidth]{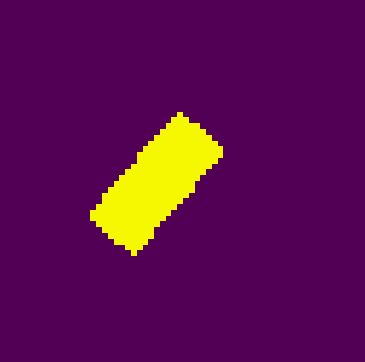}
	}
	\vspace{-0.1in}
	\caption{(a). The previous gaussian kernel heatmap assignment; (b). Our all foreground heatmap assignment.}\label{fig:gaussian_assign}
	\vspace{-0.2in}
\end{figure}

\subsection{Prediction}
Following CenterPoint~\cite{centerpoint}, we also adopt the center-manner to predict the target 3D box. Specially, in the training phase, we first generate the heatmap according to the $(x,y)$ of the 3D box center, and then compute the offset of $(x, y)$ to compensate the error from the downsample operation. For the height and orientation, we directly regress the height value of the center and $(sin\theta, cos\theta)$. However, the original heatmap in CenterPoint~\cite{centerpoint} uses Gaussian kernel locating at the center of the box, thus the number of positive samples would be not enough for the single object tracking problem since there is only one target, as shown in Fig.~\ref{fig:gaussian_assign} (a). To alleviate this problem, following SMAT~\cite{smat}, we also assign all points in the box as positive samples and generate the heatmap label as follows:
\begin{equation}
H_{p} =
\begin{cases}
1, & \text{if $p \in B$}\\
0, & \text{else}
\end{cases}       
\end{equation}
where $B$ is the 3D label box in BEV representation. In this way, we could get more positive samples during training, as shown in Fig.~\ref{fig:gaussian_assign} (b). In the inference phase, after getting the predicted heatmap and predicted offset $o$, we could compute the $x,y$ value of the center:
\begin{align}
\hat{x_c} &= (j + o_x) \times b \times v_x + x_{min}, \\ 
\hat{y_c} &= (i + o_y) \times b \times v_y + y_{min}.
\end{align} 
where $(i, j)$ is the index of the peak value in the heatmap, $(o_x, o_y)$ is the predicted offset for $(x, y)$, $b$ is the downsample stride, $(v_x, v_y)$ and $(x_{min}, y_{min})$ are the voxel size and the minimal value point cloud range of $(x, y)$ axes respectively.

\begin{table}[t]
\renewcommand\tabcolsep{1pt}
\scriptsize
\centering
\vspace{0.1in}
\caption{Performance comparison on the NuScenes dataset. The best result and the second result are marked in \textcolor{Red}{red} and \textcolor{RoyalBlue}{blue}, respectively.}~\label{tab:nuscenes}
\begin{tabular}{l|c|ccccccc}
\toprule[.05cm]
\multirow{2}*{}
& Category & Car &Ped & Truck & Bic &Bus & Trailer & Mean \\
& Frame Number &\textit{64159} &\textit{33227} &\textit{13587} &\textit{2292} &\textit{2953} & \textit{3352} &\textit{119570} \\
\hline
\hline
\multirow{7}*{Success}
& SC3D~\cite{SC3D} & 22.31 & 11.29 & 30.67 & 16.70 & 29.35 & 35.28 &20.63 \\
& P2B~\cite{P2B} &38.81 &28.39 &42.95 &26.32 &32.95 & 48.96 &36.29\\
& PTT~\cite{PTT} &41.22 &19.33 &50.23 &28.39 &\color{RoyalBlue}{43.86} &\textcolor{Red}{61.66} &36.55\\
& BAT~\cite{BAT}&40.73 &28.83 &45.34&27.17 &35.44 &52.59 &37.89\\
& V2B~\cite{v2b}&54.40 &30.10 &53.70&- &-&54.90&-\\
& C2FT~\cite{ral2} & 40.80&- & 48.40 & - &40.50&\color{RoyalBlue}{58.50}&-\\
& MLSET~\cite{ral1}&53.20 &\color{RoyalBlue}{33.20} &\textcolor{RoyalBlue}{54.30}&- &-&53.10&-\\
& $M^2$-Tracker~\cite{beyond}&\color{RoyalBlue}{55.85} &32.10 &\textcolor{Red}{57.36} &\textcolor{Red}{36.32} &\textcolor{Red}{51.39}&57.61&\color{RoyalBlue}{48.99}\\
& Ours &\textcolor{Red}{56.11} &\textcolor{Red}{37.58} &54.29 & \textcolor{RoyalBlue}{36.23}&36.31 & 48.13&\textcolor{Red}{49.66}\\
\hline
\hline
\multirow{7}*{Precision}
& SC3D~\cite{SC3D}& 21.93 & 12.65 & 27.73 & 28.12 &24.08&28.12& 20.36\\
& P2B~\cite{P2B}&43.18 &52.24 &41.59 &47.80 &27.41&40.05&45.13\\
& PTT~\cite{PTT} &45.26 &32.03 &48.56 &51.19 &\textcolor{RoyalBlue}{39.96}&\textcolor{RoyalBlue}{56.05}&42.24\\
& BAT~\cite{BAT}&43.29 &53.32 &42.58 &51.37 &28.01&44.89&45.82\\
& V2B~\cite{v2b}&59.70 &55.40 &51.10&- &- &43.70&-\\
& C2FT~\cite{ral2} & 43.80&- & 46.60 & - & 36.60&51.80&-\\
& MLSET~\cite{ral1}&58.30 &58.60 &52.50&- &-&40.90&-\\
& $M^2$-Tracker~\cite{beyond}&\textcolor{RoyalBlue}{65.09} &\textcolor{RoyalBlue}{60.92} &\textcolor{RoyalBlue}{59.54} &\textcolor{RoyalBlue}{67.50} &\textcolor{Red}{51.44}&\textcolor{Red}{58.26}&\textcolor{RoyalBlue}{62.82}\\
& Ours&\textcolor{Red}{69.07} &\textcolor{Red}{68.36} &\textcolor{Red}{60.71} & \textcolor{Red}{71.62} &36.07&55.40&\textcolor{Red}{66.77}\\
\toprule[.05cm]
\end{tabular}
\end{table}

\section{EXPERIMENTS}
\label{sec:experiments}
\subsection{Experimental Setting}
\textbf{Dataset.}
We evaluate our method on KITTI~\cite{KITTI} and NuScenes~\cite{NuScenes} datasets. For both datasets, we follow previous works~\cite{P2B,BAT,lttr,beyond} to split the training and testing sets. For NuScenes dataset, we also follow CenterPoint~\cite{centerpoint} to accumulate 10 sweeps to densify the keyframe.
	
\textbf{Implementation Details.}
Our model is implemented in Pytorch and based on the popular codebase\footnote{\url{https://github.com/open-mmlab/OpenPCDet}}, trained on RTX 3090 GPU. For feature extraction, we use dynamic pillar~\cite{dynamic} and backbone~\cite{PointPillars} which are widely used in 3D detection~\cite{second,PointPillars}. In the training phase, we train the STTracker with Adamw~\cite{adamw} optimizer with the initial learning rate of 0.003, weight decay of 0.01 for both datasets.
	
\textbf{Evaluation metric.}
One Pass Evaluation (OPE)~\cite{ope} is used to measure Success and Precision. The Success measures the 3D IoU between the predicted box and the ground-truth box, the Precision measures the AUC (area under curve) of distance between the center of two boxes from 0 to 2m. Moreover, the Mean value in each dataset is computed as follows:
\begin{equation}
V_{mean} = \frac{\sum_{n=1}^{N} V_n * F_n}{\sum_{n=1}^{N}F_n}
\end{equation}
where $V_n$ and $F_n$ represent the value and frames of each category respectively.

\begin{table}[t]
\renewcommand\tabcolsep{4pt}
\scriptsize
\begin{center}
\vspace{0.1in}
\caption{Performance comparison on the KITTI dataset. The best result and the second result are marked in \textcolor{Red}{red} and \textcolor{RoyalBlue}{blue}, respectively.}\label{tab:kitti}
		\begin{tabular}{c|c|ccccc}
			\toprule[.05cm]
			& Category & Car   & Pedestrian & Van & Cyclist &Mean\\
			& Frame Number &\textit{6424} &\textit{6088} &\textit{1248} &\textit{308} &\textit{14068} \\
			\hline
			\hline
			\multirow{15}*{Success}
			& SC3D~\cite{SC3D} &41.3 & 18.2 &40.4 &41.5 &31.2\\
			& SC3D-RPN~\cite{SC3DRPN} &36.3 & 17.9 & - &43.2 & - \\
			& P2B~\cite{P2B} &56.2 &28.7 &40.8 &32.1 &42.4\\
			& PTT~\cite{PTT} &67.8 &44.9 &43.6 &37.2 &55.1\\
			& BAT~\cite{BAT} &60.5 &42.1 &52.4 &33.7 &51.2\\
			& LTTR~\cite{lttr} &65.0 &33.2 &35.8 &66.2 &48.7\\
			& V2B~\cite{v2b} &70.5 &48.3 &50.1 &40.8 &58.4\\
            & C2FT~\cite{ral2} & 67.0 & 48.6 & 53.4 & 38.0 & 57.2\\
            & MLSET~\cite{ral1} &69.7 &50.7 &55.2 &41.0 &59.6\\
			& PTTR~\cite{pttr} &65.2 &50.9 &52.5 &65.1 &57.9\\
            & SMAT~\cite{smat}&\textcolor{RoyalBlue}{71.9} &52.1 &41.4 &61.2 &60.4\\
            & STNet~\cite{stnet} &\textcolor{Red}{72.1} &49.9
			&\textcolor{Red}{58.0} &73.5 &61.3\\
            & $M^2$-Tracker~\cite{beyond} &65.5 &\textcolor{Red}{61.5}
			&53.8 &73.2 &\textcolor{RoyalBlue}{62.9}\\
            & PCET~\cite{pcet} &68.7 &56.9 &\textcolor{RoyalBlue}{57.9} &\textcolor{Red}{75.6} &\textcolor{Red}{62.7}\\
            & STDA~\cite{stda} & 66.4 & 45.8 &-& 59.2 & -\\
			& Ours &66.5 &\textcolor{RoyalBlue}{60.4} &50.5 &\textcolor{RoyalBlue}{75.3} &62.6		\\
			\hline
			\hline
			\multirow{15}*{Precision}
			& SC3D~\cite{SC3D} &57.9 & 37.8 &47.0 &70.4 &48.5 \\
			& SC3D-RPN~\cite{SC3DRPN} &51.0 & 47.8 &- &81.2 &- \\
			& P2B~\cite{P2B} &72.8 &49.6 &48.4 &44.7 &60.0\\
			& PTT~\cite{PTT} &81.8 &72.0 &52.5 &47.3 &74.2\\
			& BAT~\cite{BAT} &77.7 &70.1 &67.0 &45.4 &72.8 \\
			& LTTR~\cite{lttr} &77.1 &56.8 &45.6 &89.9 &65.8\\
			& V2B~\cite{v2b} &81.3 &73.5 &58.0 &49.7 &75.2\\
                & C2FT~\cite{ral2} & 80.4 & 75.6 & 66.1 & 48.7 & 76.4\\
                & MLSET~\cite{ral1} &81.0 &80.0 &64.8 &49.7 &78.4\\
			& PTTR~\cite{pttr} &77.4 &81.6 &61.8 &90.5 &78.1\\
                & SMAT~\cite{smat}&\textcolor{RoyalBlue}{82.4} &81.5 &53.2 &87.3 &79.5\\
                & STNet~\cite{stnet} &\textcolor{Red}{84.0} &77.2
			&\textcolor{RoyalBlue}{70.6} &\textcolor{RoyalBlue}{93.7} &80.1\\
                & $M^2$-Tracker~\cite{beyond} &80.8 &\textcolor{RoyalBlue}{88.2} &\textcolor{Red}{70.7} &93.5 &\textcolor{Red}{83.4}\\
                & PCET~\cite{pcet} &80.1 &85.1 &66.1 &93.7 &81.3\\
            & STDA~\cite{stda} & 75.1 & 61.3& - & 72.2 & - \\ 
			& Ours &79.9 &\textcolor{Red}{89.4} &63.6 &\textcolor{Red}{93.9} &\textcolor{RoyalBlue}{82.9}\\
			\toprule[.05cm]
		\end{tabular}
	\end{center}
\end{table}

\begin{table}[t]
\renewcommand\tabcolsep{4pt}
\centering
\scriptsize
\caption{Performance comparison between ours and $M^2$-Tracker on the modified KITTI.}~\label{tab:motion_kitti}
\begin{tabular}{c|c|ccccc}\toprule[.05cm]
& Category & Car   & Pedestrian &Van &Cyclist& Mean\\
& Frame Number &\textit{1328} &\textit{1248} &\textit{255}&\textit{65}&\textit{2896}\\
\hline\hline
\multirow{2}*{Success}
& $M^2$-Tracker~\cite{beyond} &40.4&19.9 &16.4&16.6& 28.9\\
& Ours &52.2&22.8&25.6&39.3&36.9 \\
\hline\hline
\multirow{2}*{Precision}
& $M^2$-Tracker~\cite{beyond} &46.9&34.0&16.0&17.3&38.0 \\
& Ours &60.4&35.6&28.1&61.3&46.9 \\\toprule[.05cm]
\end{tabular}
\end{table}

\begin{figure*}[t]
\centering
\vspace{0.1in}
\includegraphics[width=0.9\linewidth]{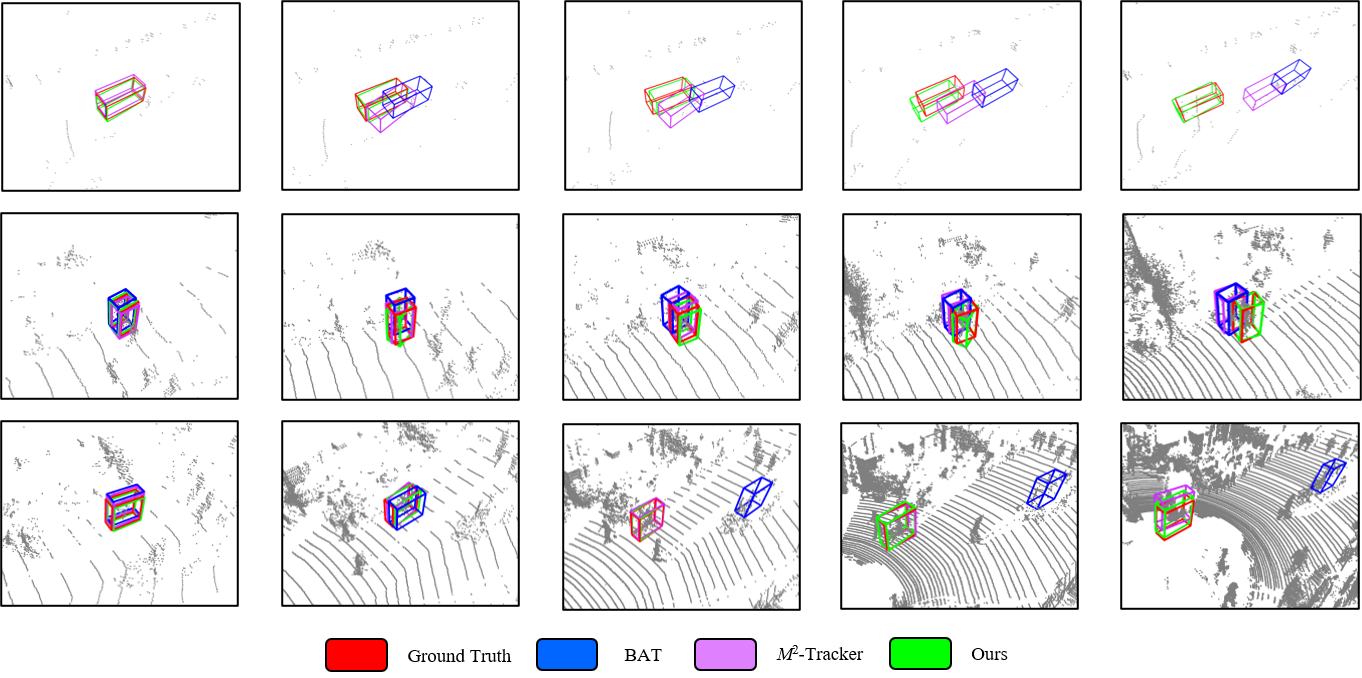}
\caption{Advantageous cases of our STTracker compared with BAT, $M^2$-Tracker on the Car, Pedestrian and Cyclist categories of KITTI Dataset. }\label{fig:kitti_results}
\end{figure*}

\begin{figure*}[t]
\centering
\includegraphics[width=0.9\linewidth]{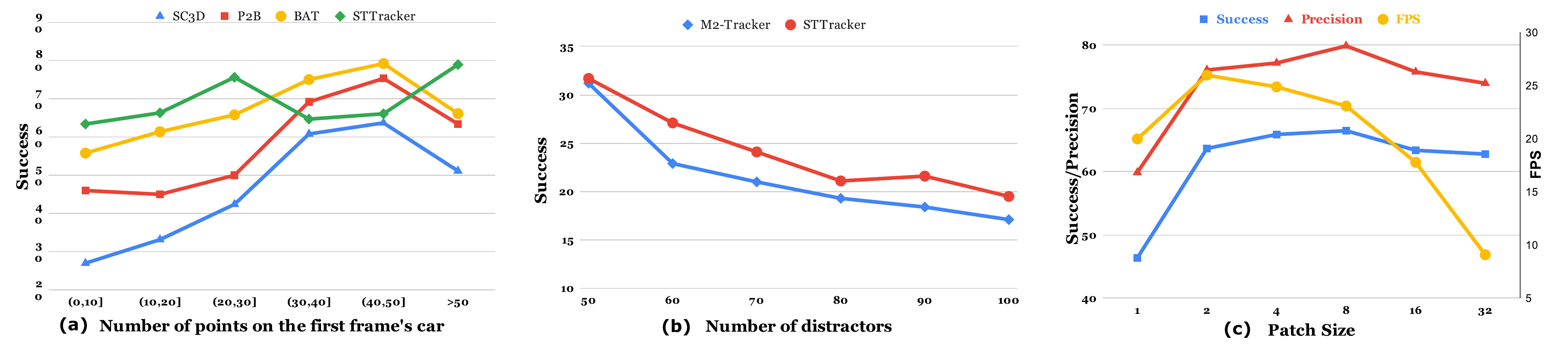}
\caption{(a) The performance of different numbers of the first-frame point cloud of the target. (b) Compared to $M^2$-Tracker under different numbers of distractors. (c) The performance with different patch sizes in spatial learning block. }\label{fig:sparsity}
\end{figure*}

\begin{table}[t]
\renewcommand\tabcolsep{4pt}
\centering
\scriptsize
\caption{Comparison of the running speeds.}\label{table:speed}
\begin{tabular}{c|cccccc}
\toprule[.05cm]
Method & SC3D & P2B   & BAT  & LTTR & V2B & PTTR      \\
FPS    & 1.8  & 45.5  & 57.0  & 22.3 & 13.0 & 51.0     \\
\hline\hline
Method  & STNet & $M^2$-Tracker & SMAT & MLSET & C2FT & STTracker \\
FPS     & 35.0  & 57.0         & 17.6 &61.0 & 50.0 & 23.6 \\
\toprule[.05cm]
\end{tabular}
\end{table}

\subsection{Comparison with State-of-the-arts}
\textbf{Results on NuScenes.} As shown in Table~\ref{tab:nuscenes}, our STTracker achieves state-of-the-art tracking performance in NuScenes dataset.
Specially, our STTracker outperforms $M^2$-Tracker by 0.26\% and 5.48\% in Success in Car and Pedestrian categories respectively, and finally has a 0.67\% improvement in the Mean. 
Meanwhile, our method shows superior performance in Precision, which exceeds $M^2$-Tracker in all categories and finally outperforms $M^2$-Tracker by 3.95\% in Mean. We believe this is due to our ability to model the motion of the target through multiple frames of input point clouds and embed this information into the measurement of similarity in the attention mechanism, thereby obtaining more accurate localization of the target. Moreover, we notice that our method performs better performance in the small-size object (Car, Pedestrain) but not that good for the large-size object (Truck, Bus, Trailer). Following the discussion in FSD~\cite{fsd}, we also believe that it is challenging for the CenterHead~\cite{centerpoint} to predict large objects since the object centers are usually empty (most of the points are on the surface of objects).

\textbf{Results on KITTI.} As shown in Table~\ref{tab:kitti}, STTracker performs competitive performance on the KITTI dataset. In terms of Success and Precision, our method only trails $M^2$-Tracker by 0.3\% and 0.5\% respectively. We believe the difference in performance between KITTI and NuScenes is due to the differing annotation frequencies of the two datasets. Specially, NuScenes is annotated at 2 Hz, while KITTI is annotated at 10 Hz, making the relative motion between frames on KITTI smaller and easier to estimate, thus it is more advantageous for $M^2$-Tracker which directly predicts relative motion. To verify this assumption, we further compare our method and $M^2$-Tracker~\cite{beyond} in a modified KITTI dataset. To have the same annotation frequency as NuScenes dataset, \textit{i.e.} 2Hz, we only select the tracklets which have more than 5 frames for training and testing, and sample one frame as a valid frame every 5 frames. We train our model and $M^2$-Tracker on the new dataset. The training settings of $M^2$-Tracker is following their public setting\footnote{\url{https://github.com/Ghostish/Open3DSOT/tree/main/cfgs}}. Shown in Table.~\ref{tab:motion_kitti}, our method shows better performance than $M^2$-Tracker on the modified KITTI dataset, verifying our assumption that our method could have a better performance in large motion tracking scenes.
Meanwhile, for the previous similarity-based methods~\cite{BAT,pttr,v2b,lttr,PTT,smat,pcet}, although they had good performances in the Mean, they usually performed worse in Pedestrian category. We believe that the size of Pedestrian is small thus limiting the similarity-based methods. However, our method outperforms PCET, which had the best performance among similarity-based methods in Pedestrian, by 3.5\% and 4.3\% in Success and Precision respectively. The results show that by learning the spatio-temporal information, our STTracker could achieve better performance. Moreover, our method also outperforms STDA which also use spatio-temporal information
Additionally, our STTracker achieves \textbf{23.6 FPS} running speed shown in Table.~\ref{table:speed}, and we also visualize the tracking results in Fig~\ref{fig:kitti_results}.

\begin{table}[t]
\renewcommand\tabcolsep{2pt}
\scriptsize
\vspace{0.1in}
\caption{Ablation of our components. MF and FG stands for multi frames and foreground assignment respectively.}\label{tab:ablation}
\begin{adjustbox}{center}
\begin{tabular}{c|ccc|cc|cc}\toprule[.05cm]
\multirow{2}{*}{} &
\multirow{2}{*}{MF} &\multirow{2}{*}{STLM} & \multirow{2}{*}{FG}&			
\multicolumn{2}{c|}{3D} & \multicolumn{2}{c}{BEV} \\
&&&& Success & Precision & Success & Precision\\ \hline 
\midrule
A1&   & & &59.6&70.0&66.0&71.2 \\
A2&\checkmark  &  &  &51.5 \reddown\color{red}{8.1\%}&61.4 \reddown{\color{red}8.6\%}&57.4 \reddown{\color{red}8.6\%}&62.7 \reddown\color{red}{8.5\%}\\
A3&\checkmark   &\checkmark& &63.3 \greenup\color{Green}{3.7\%}&74.4 \greenup\color{Green}{4.4\%}& 70.7 \greenup\color{Green}{4.7\%}&76.0 \greenup\color{Green}{4.8\%}\\
A4&\checkmark &  \checkmark &\checkmark &66.5 \greenup\color{Green}{6.9\%}&79.9 \greenup\color{Green}{9.9\%}&74.6 \greenup\color{Green}{8.6\%}&82.1 \greenup\color{Green}{10.9\%}\\
\toprule[.05cm]
\end{tabular}
\end{adjustbox}
\end{table}

\begin{table}[t]
\renewcommand\tabcolsep{5pt}
\scriptsize
\vspace{0.1in}
\caption{Ablation of different input frames.}\label{tab:input_frames}
\begin{adjustbox}{center}
\begin{tabular}{c|c|cccc}\toprule[.05cm]
& Input Time ID  & Success & Precision &FPS \\ \hline \midrule
I1 &(t, t-1) & 64.0    & 75.8 &  \textbf{36.8}  \\
I2 &(t, t-2) & 64.2    & 75.9& 36.8    \\
I3 &(t, t-1, t-2)  & 65.8   & 78.8&   28.6  \\
I4 &(t, t-1, t-2, t-3)  & \textbf{66.5}    & \textbf{79.9} & 23.6 \\
I5 &(t, t-1, t-3, t-5)  & 59.8    & 72.1  & 23.6 \\
I6 &(t, t-2, t-3, t-4)  & 61.4    & 73.1  &23.6 \\
I7 &(t, t-2, t-4, t-6)  & 63.1    & 77.3  &23.6\\
I8 &(t, t-1, t-2, t-3, t-4)   & 64.8   & 75.7 & 20.0   \\ 
I9 &(t, t-1, t-2, t-3, t-4, t-5)   & 62.8   & 76.2  & 17.4  \\ \toprule[.05cm]
\end{tabular}
\end{adjustbox}
\end{table}

\textbf{Robust to Sparsity.} To explore the robustness to the sparse point cloud, we classify the car tracking sequences in KITTI into different levels according to the number of point clouds in the first frame. The proposed STTracker is then evaluated on these sequences. As shown in Fig~\ref{fig:sparsity} (a), STTracker outperforms the other methods in tracking sparse targets with fewer than 30 points in the first frame. Meanwhile, our method achieves similar tracking performance at different sparse levels, verifying the robustness of sparsity.

\textbf{Robust to distractors.} To further explore our robustness to distractors, we compare our method with $M^2$-Tracker~\cite{beyond} under different numbers of distractors. We randomly add $K$ car instances to the testing scenes of KITTI, and then evaluate their pretrained model\footnote{\url{https://github.com/Ghostish/Open3DSOT/blob/main/pretrained_models/mmtrack_kitti_car.ckpt}} and our trained models using these synthesis sequences. As shown in Fig.~\ref{fig:sparsity} (b), although our method and $M^2$-Tracker~\cite{beyond} both have a large performance decline, our method still achieves better performance than $M^2$-Tracker, verifying our robustness to distractors.

\subsection{Ablation Study}
In this section, we conduct comprehensive experiments to validate the design of STTracker. All experiments are conducted on the Car category of the KITTI dataset.

\textbf{Multi-frames Input.}
We first input different number and different timestamps to STTracker, as shown in Table.~\ref{tab:input_frames}. Compared to the other settings, I4 achieves the best performance. Specially, I1 and I2 only input a single frame thus could not learn enough temporal information. Meanwhile, compared to I4, I8 and I9 have more frames but achieve worse performance. We believe that too many input frames would introduce too much cumulative error during tracking. Meanwhile, I5, I6 and I7 have the same input length as I4, but still perform worse than I4. We believe they also suffer from the cumulative error due to their longer input range. Notice that different input settings need to be retrained separately.

\begin{table}[t]
\centering
\scriptsize
\vspace{0.1in}
\caption{Ablation of spatial learning block.}\label{tab:slb}
\begin{tabular}{cc|cc|cc}\toprule[.05cm]
    & \multirow{2}{*}{Method} &\multicolumn{2}{c|}{3D} & \multicolumn{2}{c}{BEV} \\
	& & Success & Precision & Success & Precision\\ \hline \midrule
S1 & dot &  65.0   &   76.2  & 71.9   &77.7  \\
S2 & w/o Mask  & 62.7 &   75.1   & 71.3  & 77.2            \\
S3 & w/o BoxConv   &  64.9   &  77.8   & 72.3    & 79.6    \\ 
S4 & Conv-Patch & 65.3 &   78.9   & 72.1  & 80.5  \\\toprule[.05cm] 
\end{tabular}
\vspace{-0.1in}
\end{table}

\textbf{Spatial Learning Block.}
The study of spatial learning block includes the $MaskFusion$, the patch transformation and the patch size, as shown in Table.~\ref{tab:slb}. S1 represents replacing the $MaskConv$ and ``add'' operation in $MaskFusion$ with dot multiplication, the performance is lower than our method by 1.5\% and 3.7\% in Success and Precision respectively. We believe that the simple dot multiplication eliminates the context of surrounding information, thus breaking the spatial information of the target. 
Not surprisingly, without the mask to distinguish the target and background (S2), the method has a large decline, 3.8\%$\downarrow$ and 4.8\%$\downarrow$ in Success and Precision respectively. S3 also shows the importance of $BoxConv$ in mask fusion. In S4, we change the order of patch transformation and the following Conv2D layer, resulting in 1.2\%$\downarrow$ and 1.0\%$\downarrow$ in Success and Precision respectively. We believe that compared to the "Conv-Patch" order, our "Patch-Conv" order could better aggregate the features within the patch, thus benefiting the subsequent temporal learning block.
Moreover, we also show the performances of different patch sizes, as shown in Fig.~\ref{fig:sparsity} (c). We believe that a small patch size has not enough receptive field for comparison, while a large patch size covers too many areas thus only getting a coarse comparison.

\begin{table}[t]
\centering
\scriptsize
\vspace{0.1in}
\caption{Ablation of temporal learning block.}\label{tab:tlb}
\begin{tabular}{cc|cc|cc}\toprule[.05cm]
& \multirow{2}{*}{Method} &\multicolumn{2}{c|}{3D} & \multicolumn{2}{c}{BEV} \\
& & Success & Precision & Success & Precision\\ \hline \midrule
T1 & w/o $P_t$ &65.8    & 77.8  &  73.3  &79.7\\
T2 & w/o $F_t$ & 56.8    & 69.4 &64.7 & 71.4 \\
T3 & dense attention & 61.4  & 76.3  & 68.8     &78.1 \\
T4 & w/. PE & 65.2 & 77.5   & 72.5   & 79.3   \\\toprule[.05cm]    
\end{tabular}
\vspace{-0.1in}
\end{table}

\textbf{Temporal Learning Block.}
We further try different components in temporal learning block, as shown in Table.~\ref{tab:tlb}. The results of T1 and T2 verify the importance of current feature in fusion. Meanwhile, instead of using sparse attention, T3 uses dense attention and the performance drops 5.1\% and 3.6\% in 3D Success and Precision respectively. The results show that there is no need to compare all location features in fusion for different frame features. Lastly, because the point feature already includes the 3D information and extra time feature, adding positional embedding (T4) does not improve the performance.

\textbf{Ablation Experiments.}
Finally, we conduct ablation experiments on the components of our method. Table.~\ref{tab:ablation} shows the results. A1 is the baseline model which only inputs two frames. A2 shows that directly concatenating the multi-frame features could not bring improvement but a huge decrease, as analyzed in Sec.~\ref{sec:stlearning}. Compared to A2, A3 shows great improvement, 11.8\%$\uparrow$ and 13.0\%$\uparrow$ in 3D Success and Precision, which verifies the effeteness of our proposed STLM. Finally, by using the foreground heatmap assignment, A4 achieves the best performance.

\section{CONCLUSIONS}
\label{sec:conclusions}
In this paper, we present STTracker, a multi-frame similarity-based tracking framework to track 3D object with point cloud. 
We propose a spatio-temporal learning module to fuse multi-frame features and fully exploit the spatio-temporal information of 3D target. 
The comprehensive experiments show the effectiveness of our method. 
Meanwhile, We notice that our method does not have obvious advantages in large-size objects or high-frequency scenes, and too much input frames also leads to the performance decline. Therefore, we would like to solve these problems in future works.

\bibliographystyle{IEEEtran}
\bibliography{stt}

\begin{thebibliography}{10}
\providecommand{\url}[1]{#1}
\csname url@samestyle\endcsname
\providecommand{\newblock}{\relax}
\providecommand{\bibinfo}[2]{#2}
\providecommand{\BIBentrySTDinterwordspacing}{\spaceskip=0pt\relax}
\providecommand{\BIBentryALTinterwordstretchfactor}{4}
\providecommand{\BIBentryALTinterwordspacing}{\spaceskip=\fontdimen2\font plus
\BIBentryALTinterwordstretchfactor\fontdimen3\font minus
  \fontdimen4\font\relax}
\providecommand{\BIBforeignlanguage}[2]{{%
\expandafter\ifx\csname l@#1\endcsname\relax
\typeout{** WARNING: IEEEtran.bst: No hyphenation pattern has been}%
\typeout{** loaded for the language `#1'. Using the pattern for}%
\typeout{** the default language instead.}%
\else
\language=\csname l@#1\endcsname
\fi
#2}}
\providecommand{\BIBdecl}{\relax}
\BIBdecl

\bibitem{KITTI}
A.~Geiger, P.~Lenz, and R.~Urtasun, ``Are we ready for autonomous driving? the
  kitti vision benchmark suite,'' in \emph{CVPR}, 2012, pp. 3354--3361.

\bibitem{NuScenes}
H.~Caesar, V.~Bankiti, A.~H. Lang, S.~Vora, V.~E. Liong, Q.~Xu, A.~Krishnan,
  Y.~Pan, G.~Baldan, and O.~Beijbom, ``nuscenes: A multimodal dataset for
  autonomous driving,'' in \emph{CVPR}, 2020, pp. 11\,621--11\,631.

\bibitem{votenet}
C.~R. Qi, O.~Litany, K.~He, and L.~J. Guibas, ``Deep hough voting for 3d object
  detection in point clouds,'' in \emph{ICCV}, 2019, pp. 9277--9286.

\bibitem{SC3D}
S.~Giancola, J.~Zarzar, and B.~Ghanem, ``Leveraging shape completion for 3d
  siamese tracking,'' in \emph{CVPR}, 2019, pp. 1359--1368.

\bibitem{SC3DRPN}
J.~Zarzar, S.~Giancola, and B.~Ghanem, ``Efficient bird eye view proposals for
  3d siamese tracking,'' \emph{ArXiv}, vol. abs/1903.10168, 2019.

\bibitem{P2B}
H.~Qi, C.~Feng, Z.~Cao, F.~Zhao, and Y.~Xiao, ``P2b: Point-to-box network for
  3d object tracking in point clouds,'' in \emph{CVPR}, 2020, pp. 6329--6338.

\bibitem{BAT}
C.~Zheng, X.~Yan, J.~Gao, W.~Zhao, W.~Zhang, Z.~Li, and S.~Cui, ``Box-aware
  feature enhancement for single object tracking on point clouds,'' in
  \emph{ICCV}, 2021, pp. 13\,199--13\,208.

\bibitem{PTT}
J.~Shan, S.~Zhou, Z.~Fang, and Y.~Cui, ``Ptt: Point-track-transformer module
  for 3d single object tracking in point clouds,'' in \emph{IROS}, 2021, pp.
  1310--1316.

\bibitem{lttr}
Y.~Cui, Z.~Fang, J.~Shan, Z.~Gu, and S.~Zhou, ``3d object tracking with
  transformer,'' in \emph{32nd BMVC}, 2021, p. 317.

\bibitem{v2b}
L.~Hui, L.~Wang, M.~Cheng, J.~Xie, and J.~Yang, ``3d siamese voxel-to-bev
  tracker for sparse point clouds,'' in \emph{NeurIPS}, vol.~34, 2021, pp.
  28\,714--28\,727.

\bibitem{beyond}
C.~Zheng, X.~Yan, H.~Zhang, B.~Wang, S.~Cheng, S.~Cui, and Z.~Li, ``Beyond 3d
  siamese tracking: A motion-centric paradigm for 3d single object tracking in
  point clouds,'' in \emph{CVPR}, 2022, pp. 8111--8120.

\bibitem{pttr}
C.~Zhou, Z.~Luo, Y.~Luo, T.~Liu, L.~Pan, Z.~Cai, H.~Zhao, and S.~Lu, ``Pttr:
  Relational 3d point cloud object tracking with transformer,'' in \emph{CVPR},
  2022, pp. 8531--8540.

\bibitem{stnet}
L.~Hui, L.~Wang, L.~Tang, K.~Lan, J.~Xie, and J.~Yang, ``3d siamese transformer
  network for single object tracking on point clouds,'' vol. abs/2207.11995,
  2022.

\bibitem{mgtanet}
J.~Koh, J.~Lee, Y.~Lee, J.~Kim, and J.~W. Choi, ``Mgtanet: Encoding sequential
  lidar points using long short-term motion-guided temporal attention for 3d
  object detection,'' 2022.

\bibitem{VIT}
A.~Dosovitskiy, L.~Beyer, A.~Kolesnikov, D.~Weissenborn, X.~Zhai,
  T.~Unterthiner, M.~Dehghani, M.~Minderer, G.~Heigold, S.~Gelly, J.~Uszkoreit,
  and N.~Houlsby, ``An image is worth 16x16 words: Transformers for image
  recognition at scale,'' in \emph{9th ICLR}, 2021.

\bibitem{pvt}
W.~Wang, E.~Xie, X.~Li, D.-P. Fan, K.~Song, D.~Liang, T.~Lu, P.~Luo, and
  L.~Shao, ``Pyramid vision transformer: A versatile backbone for dense
  prediction without convolutions,'' in \emph{ICCV}, 2021, pp. 568--578.

\bibitem{TNT}
K.~Han, A.~Xiao, E.~Wu, J.~Guo, C.~Xu, and Y.~Wang, ``Transformer in
  transformer,'' in \emph{NeurIPS}, vol.~34, 2021, pp. 15\,908--15\,919.

\bibitem{swin}
Z.~Liu, Y.~Lin, Y.~Cao, H.~Hu, Y.~Wei, Z.~Zhang, S.~Lin, and B.~Guo, ``Swin
  transformer: Hierarchical vision transformer using shifted windows,'' in
  \emph{ICCV}, 2021, pp. 10\,012--10\,022.

\bibitem{adamw}
I.~Loshchilov and F.~Hutter, ``Decoupled weight decay regularization,'' in
  \emph{7th ICLR}, 2019.

\bibitem{faf}
W.~Luo, B.~Yang, and R.~Urtasun, ``Fast and furious: Real time end-to-end 3d
  detection, tracking and motion forecasting with a single convolutional net,''
  in \emph{CVPR}, 2018.

\bibitem{StarNet}
J.~Ngiam, B.~Caine, W.~Han, B.~Yang, Y.~Chai, P.~Sun, Y.~Zhou, X.~Yi,
  O.~Alsharif, P.~Nguyen, Z.~Chen, J.~Shlens, and V.~Vasudevan, ``Starnet:
  Targeted computation for object detection in point clouds,'' 2019.

\bibitem{STINet}
Z.~Zhang, J.~Gao, J.~Mao, Y.~Liu, D.~Anguelov, and C.~Li, ``Stinet:
  Spatio-temporal-interactive network for pedestrian detection and trajectory
  prediction,'' in \emph{CVPR}, 2020.

\bibitem{st3dcnn}
B.~Mersch, X.~Chen, J.~Behley, and C.~Stachniss, ``Self-supervised point cloud
  prediction using 3d spatio-temporal convolutional networks,'' in \emph{5th
  ICRL}, 2021.

\bibitem{MVFuseNet}
A.~Laddha, S.~Gautam, S.~Palombo, S.~Pandey, and C.~Vallespi-Gonzalez,
  ``Mvfusenet: Improving end-to-end object detection and motion forecasting
  through multi-view fusion of lidar data,'' \emph{CVPRW}, 2021.

\bibitem{PointNet}
C.~R. Qi, H.~Su, K.~Mo, and L.~J. Guibas, ``Pointnet: Deep learning on point
  sets for 3d classification and segmentation,'' in \emph{CVPR}, 2017, pp.
  652--660.

\bibitem{PointNet++}
C.~R. Qi, L.~Yi, H.~Su, and L.~J. Guibas, ``Pointnet++: Deep hierarchical
  feature learning on point sets in a metric space,'' in \emph{NeurIPS}, 2017,
  p. 5105–5114.

\bibitem{second}
Y.~Yan, Y.~Mao, and B.~Li, ``Second: Sparsely embedded convolutional
  detection,'' in \emph{Sensors}, vol.~18, no.~10, 2018, p. 3337.

\bibitem{PointPillars}
A.~H. Lang, S.~Vora, H.~Caesar, L.~Zhou, J.~Yang, and O.~Beijbom,
  ``Pointpillars: Fast encoders for object detection from point clouds,'' in
  \emph{CVPR}, 2019, pp. 12\,697--12\,705.

\bibitem{dynamic}
Y.~Zhou, P.~Sun, Y.~Zhang, D.~Anguelov, J.~Gao, T.~Ouyang, J.~Guo, J.~Ngiam,
  and V.~Vasudevan, ``End-to-end multi-view fusion for 3d object detection in
  lidar point clouds,'' in \emph{Conference on Robot Learning}.\hskip 1em plus
  0.5em minus 0.4em\relax PMLR, 2020, pp. 923--932.

\bibitem{VoxelNet}
Y.~Zhou and O.~Tuzel, ``Voxelnet: End-to-end learning for point cloud based 3d
  object detection,'' in \emph{CVPR}, 2018, pp. 4490--4499.

\bibitem{pv_rcnn}
S.~Shi, C.~Guo, L.~Jiang, Z.~Wang, J.~Shi, X.~Wang, and H.~Li, ``Pv-rcnn:
  Point-voxel feature set abstraction for 3d object detection,'' in
  \emph{CVPR}, 2020, pp. 10\,529--10\,538.

\bibitem{ope}
Y.~Wu, J.~Lim, and M.-H. Yang, ``Online object tracking: A benchmark,'' in
  \emph{CVPR}, 2013, pp. 2411--2418.

\bibitem{centerpoint}
T.~Yin, X.~Zhou, and P.~Krahenbuhl, ``Center-based 3d object detection and
  tracking,'' \emph{2021 IEEE/CVF Conference on Computer Vision and Pattern
  Recognition (CVPR)}, pp. 11\,784--11\,793, 2021.

\bibitem{Transformer}
A.~Vaswani, N.~Shazeer, N.~Parmar, J.~Uszkoreit, L.~Jones, A.~N. Gomez, L.~u.
  Kaiser, and I.~Polosukhin, ``Attention is all you need,'' in \emph{Advances
  in Neural Information Processing Systems}, vol.~30.\hskip 1em plus 0.5em
  minus 0.4em\relax Curran Associates, Inc., 2017.

\bibitem{DeformableDETR}
X.~Zhu, W.~Su, L.~Lu, B.~Li, X.~Wang, and J.~Dai, ``Deformable detr: Deformable
  transformers for end-to-end object detection,'' in \emph{ICLR}, 2021.

\bibitem{focal}
T.-Y. Lin, P.~Goyal, R.~Girshick, K.~He, and P.~Dollar, ``Focal loss for dense
  object detection,'' in \emph{ICCV}, Oct 2017.

\bibitem{smat}
Y.~Cui, J.~Shan, Z.~Gu, Z.~Li, and Z.~Fang, ``Exploiting more information in
  sparse point cloud for 3d single object tracking,'' in \emph{IEEE Robotics
  and Automation Letters}, vol.~7, no.~4, 2022, pp. 11\,926--11\,933.

\bibitem{ral1}
Q.~Wu, C.~Sun, and J.~Wang, ``Multi-level structure-enhanced network for 3d
  single object tracking in sparse point clouds,'' \emph{IEEE Robotics and
  Automation Letters}, vol.~8, no.~1, pp. 9--16, 2023.

\bibitem{ral2}
B.~Fan, K.~Wang, H.~Zhang, and J.~Tian, ``Accurate 3d single object tracker
  with local-to-global feature refinement,'' \emph{IEEE Robotics and Automation
  Letters}, vol.~7, no.~4, pp. 12\,211--12\,218, 2022.

\bibitem{pcet}
P.~Wang, L.~Ren, S.~Wu, J.~Yang, E.~Yu, H.~Yu, and X.~Li, ``Implicit and
  efficient point cloud completion for 3d single object tracking,'' \emph{IEEE
  Robotics and Automation Letters}, vol.~8, no.~4, pp. 1935--1942, 2023.

\bibitem{pf_track}
Z.~Pang, J.~Li, P.~Tokmakov, D.~Chen, S.~Zagoruyko, and Y.-X. Wang, ``Standing
  between past and future: Spatio-temporal modeling for multi-camera 3d
  multi-object tracking,'' in \emph{CVPR}, 2023.

\bibitem{spot}
C.~Stearns, D.~Rempe, J.~Li, R.~Ambrus, S.~Zakharov, V.~Guizilini, Y.~Yang, and
  L.~J. Guibas, ``Spot: Spatiotemporal modeling for 3d object tracking,'' in
  \emph{ECCV}, 2022.

\bibitem{stda}
Y.~Zhang, H.~Niu, Y.~Guo, and W.~He, ``3d single-object tracking with
  spatial-temporal data association,'' in \emph{2022 IEEE/RSJ International
  Conference on Intelligent Robots and Systems (IROS)}, 2022, pp. 264--269.

\bibitem{fsd}
L.~Fan, F.~Wang, N.~Wang, and Z.~Zhang, ``{Fully Sparse 3D Object Detection},''
  in \emph{NeurIPS}, 2022.

\bibitem{detr}
\BIBentryALTinterwordspacing
N.~Carion, F.~Massa, G.~Synnaeve, N.~Usunier, A.~Kirillov, and S.~Zagoruyko,
  ``End-to-end object detection with transformers,'' \emph{ECCV}, p. 213–229,
  2020. [Online]. Available:
  \url{http://dx.doi.org/10.1007/978-3-030-58452-8_13}
\BIBentrySTDinterwordspacing

\end{thebibliography}

\end{document}